%% file: main.tex
\documentclass[sigconf]{acmart}
\AtBeginDocument{%
  }

\setcopyright{acmlicensed}
\copyrightyear{2026}
\acmYear{2026}
\setcopyright{cc}
\setcctype{by-nc-nd}
\acmConference[MM '26]{Proceedings of the 34th ACM International Conference on Multimedia}{November 10--14, 2026}{Rio de Janeiro, Brazil}
\acmBooktitle{Proceedings of the 34th ACM International Conference on Multimedia (MM '26), November 10--14, 2026, Rio de Janeiro, Brazil}
\acmDOI{10.1145/3767308.3835641}
\acmISBN{979-8-4007-2213-4/2026/11}

\usepackage{enumitem}
\usepackage{multirow}

\begin{document}

\title{GHOST: Geometry-Constrained Hybrid Diffusion for Open-Vocabulary 3D Scene Graph Generation}

\author{Changqun Feng}
\affiliation{%
  \institution{University of Electronic Science and Technology of China}
  \city{Chengdu}
  \state{Sichuan}
  \country{China}}
\email{fchangqun@gmail.com}

\author{Wangxiandi Yin}
\affiliation{%
  \institution{University of Electronic Science and Technology of China}
  \city{Chengdu}
  \state{Sichuan}
  \country{China}}
\email{yinwangxiandi@gmail.com}

\author{Xin Hu}
\affiliation{%
  \institution{University of Electronic Science and Technology of China}
  \city{Chengdu}
  \state{Sichuan}
  \country{China}}
\email{xh1m22@std.uestc.edu.cn}

\author{Ruiting Dai}
\affiliation{%
  \institution{University of Electronic Science and Technology of China}
  \city{Chengdu}
  \state{Sichuan}
  \country{China}}
\email{rtdai@uestc.edu.cn}

\author{Feiyang Yang}
\affiliation{%
  \institution{Jilin University}
  \city{Changchun}
  \state{Jilin}
  \country{China}}
\email{yangfy21@mails.jlu.edu.cn}

\author{Jielei Wang}
\affiliation{%
  \institution{University of Electronic Science and Technology of China}
  \city{Chengdu}
  \state{Sichuan}
  \country{China}}
\email{jieleiwang\_uestc@163.com}

\author{Tao He}
\authornote{Corresponding author.}
\affiliation{%
  \institution{University of Electronic Science and Technology of China}
  \city{Chengdu}
  \state{Sichuan}
  \country{China}}
\email{tao.he01@hotmail.com}

\renewcommand{\shortauthors}{Changqun Feng et al.}

\begin{abstract}
3D scene graph generation requires understanding both object semantics and metric spatial relations. However, existing open-vocabulary methods often struggle with spatial predicates because language similarity reflects semantic co-occurrence rather than explicit geometry. We propose \textbf{GHOST}, a \underline{G}eometry-constrained \underline{H}ybrid diffusion framework for \underline{O}pen-vocabulary 3D \underline{S}cene graph genera\underline{T}ion. GHOST jointly models continuous node geometry and discrete edge predicates, following a geometry-to-constraint-to-predicate denoising process to ensure geometrically consistent relation prediction. It introduces language-grounded constraint induction, which maps predicate embeddings to differentiable geometric constraints that guide both layout refinement and predicate reweighting. A Geometry-Topology Collaborative Denoiser further couples node and edge reasoning through noise-adaptive interaction and orientation-aware modeling. Experiments on 3DSSG demonstrate state-of-the-art performance in both closed-set and open-vocabulary settings, including over 17-point gains over RelationField on spatial predicates.

\end{abstract}

\ccsdesc[500]{Computing methodologies~Computer vision}

\keywords{Diffusion Models; 3D Scene Graph Generation; Spatial Reasoning}

\maketitle

\input{sec/1_introduction}

\input{sec/2_relatedwork-ht}

\input{sec/3_preliminary}
\input{sec/4_method-ht}
\input{sec/5_experiments}

\input{sec/6_Conclusions}
\bibliographystyle{ACM-Reference-Format}
\bibliography{acmart} %

\end{document}

%% file: sec/1_introduction.tex
\section{Introduction}
\label{sec:intro}

\begin{figure}[t]
  \centering
  \includegraphics[width=0.47\textwidth]{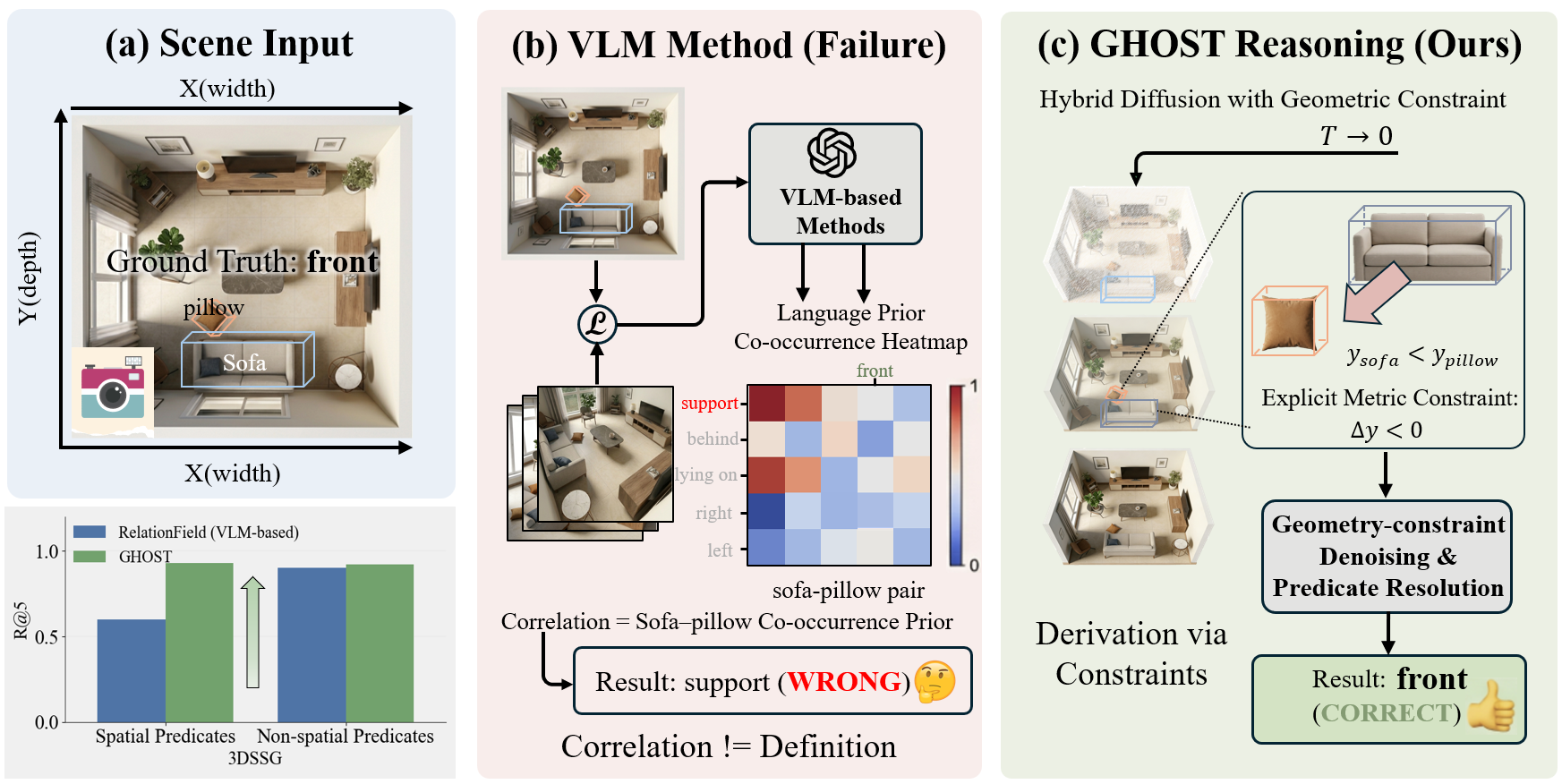}
  \caption{
  VLM-based open-vocabulary 3DSGG methods perform about 30 points worse on spatial predicates than on non-spatial ones (a), because they rely on language co-occurrence priors rather than true 3D layout, e.g., a model may predict “pillow on sofa” simply because the relation is common, while ignoring the actual scene layout (b). GHOST replaces such priors with explicit geometric constraints, substantially narrowing this gap (c).
  }  
  \label{fig:intro}
\end{figure}

Understanding a 3D scene requires not only recognizing which objects are present, but also reasoning about how they spatially relate. Is the pillow \emph{on} or \emph{in front of} the sofa? (Fig.~\ref{fig:intro}(a)) Answering such questions precisely is the central challenge of 3D Scene Graph Generation (3DSGG)~\cite{koch2024open3dsg, chen2024clip,DBLP:conf/mm/LiZ0XLQ24,feng2026geode,he2026lifelong}, which represents a scene as a structured graph where nodes are localized objects and directed edges encode their spatial, comparative, and functional relations.

The field has progressed from closed-set models~\cite{koch2024lang3dsg, feng20233d, he2023toward}, which operate on fixed object and predicate vocabularies, to open-vocabulary (OV) methods~\cite{koch2024open3dsg,  Yan_2025_CVPR, Liu_2025_CVPR,li2024instant3d, DBLP:conf/aaai/LiZDXQ25,koch2025relationfield} that leverage vision--language model (VLM) priors to handle unseen categories.
Although this transition greatly improves semantic coverage, it also exposes a key limitation: VLMs encode spatial relations from image--text co-occurrence rather than verifiable geometric constraints over metric 3D coordinates~\cite{Ho2020, cheng2024spatialrgpt,hu2025spade}.
As a result, spatial reasoning remains fragile (Fig. \ref{fig:intro}(b)).
In our controlled diagnostic (Sec.~\ref{subsec:Comparison}), which evaluates spatial and non-spatial predicates separately, leading VLM-based pipelines~\cite{ koch2025relationfield} score about $30$ R@5 points lower on spatial predicates than on non-spatial ones (the bottom of Fig.~\ref{fig:intro}(a)), with failures concentrated in depth ambiguities and symmetric configurations.

A natural remedy is to inject 3D cues, such as bounding-box coordinates or point-cloud features, into the VLM pipeline.
In practice, this often fails and may even worsen generalization.
When geometry is simply appended to visual or textual features~\cite{yang2026cam,yin2026tical}, the model must weigh it against strong pre-trained language priors.
Because indoor spatial statistics are highly structured, e.g., cabinets usually appear above counters and chairs in front of desks, coordinate features easily become shortcuts for predicate prediction rather than inputs to true geometric reasoning~\cite{geirhos2020shortcut, tang2020unbiased,li2026fixed}.
The model thus learns correlations between coordinate patterns and predicates, but degrades when these correlations shift~\cite{zhang2021exploiting, wu2021scenegraphfusion}.
The problem is therefore not the absence of geometric information, but the absence of the right inductive structure.
A spatial predicate is not merely associated with geometry; it is defined by it.
For example, ``left of'' is true if and only if the subject's $x$-coordinate is smaller than the object's.

Given the above analysis, one might propose a simpler fix, predicting predicates in a single forward pass and then applying a geometric consistency check.
We argue that this two-stage strategy is insufficient in principle, and confirm this empirically in Sec.~\ref{sec:exp:oracle}.
Its limitation is that one-shot prediction forces an early commitment before the geometric layout is fully resolved.
For instance, when two objects have nearly identical depths, a one-shot model cannot represent uncertainty over their ordering and must still choose between ``in front of'' and ``behind.''
Such decisions are therefore driven by training-set frequency rather than the true scene geometry.
Post-hoc verification may detect the inconsistency, but it cannot reliably recover the correct predicate once a hard prediction has been made.
This calls for a generative process in which layout and relation estimates remain soft and mutually revisable until sufficient geometric evidence has been accumulated.

Building on the above insight, we introduce \textbf{GHOST}, a \underline
{G}eometry-constrained \underline{H}ybrid diffusion framework for \underline{O}pen-vocabulary 3D \underline{S}cene graph genera\underline{T}ion.
GHOST maintains a joint distribution over continuous node geometry and discrete edge predicates through a hybrid diffusion process, resolving both iteratively across $T$ reverse steps.
\emph{First}, within each reverse step we enforce a strict {geometric annealing schedule} in which geometry is denoised, then projected onto a constraint manifold, and finally predicates are sampled conditioned on the constraint-consistent layout.
This intra-step ordering encodes the definitional dependence of spatial predicates on metric coordinates as a hard structural prior at every step of the reverse process.
\emph{Second}, we propose {language-grounded constraint guidance}, a mechanism that maps any predicate's CLIP embedding, including predicates unseen during training, to a differentiable constraint function composed over a shared basis of geometric primitives (axis displacements, proximity, containment, relative size).
The resulting constraint energy steers both geometry via gradient correction and predicate probabilities via distribution reweighting, without requiring a closed vocabulary of spatial predicates.

Finally, \emph{Geometry-Topology Collaborative Denoiser} (GTCD) jointly refines nodes and edges via noise-adaptive gating, edge-conditioned attention that feeds relational cues back to geometry, and an orien-tation-aware edge decomposition tailored to asymmetric predicates such as “left of” and “above.” As a result, spatial predicates are inferred from recovered geometry under explicit constraints rather than directly classified from visual–linguistic features. This progressive co-refinement yields consistent gains over one-shot baselines, particularly for near-depth relations, highlighting the value of joint uncertainty resolution.

In summary, we make the following contributions.
\begin{itemize}[leftmargin=*,noitemsep,topsep=2pt]
  \item We analyze why VLM-based 3DSGG often fails on spatial predicates, and show that the issue mainly comes from co-occurrence bias and one-shot prediction, which post-hoc geometric correction cannot fully fix.

  \item We propose \textbf{GHOST}, a geometry-constrained  hybrid continuous-discrete diffusion framework for  3DSGG. Its semantic constraint induction maps predicates to differentiable geometric constraints, enabling spatial reasoning.

  \item We design  \emph{Geometry-Topology Collaborative Denoiser} (GTCD), which  couples geometry and relations via noise-adaptive interaction and geometry-aware edge modeling.

  \item GHOST achieves state-of-the-art results on 3DSSG under both closed-set and open-vocabulary settings, with particularly strong gains on spatial predicates.
\end{itemize}

%% file: sec/2_relatedwork-ht.tex
\section{Related Work}
\label{sec:related}

\noindent\textbf{3D Scene Graph Generation.}
Early 3D scene graph generation methods assume closed-set object and predicate vocabularies~\cite{wald2020learning, wu2023incremental,DBLP:journals/tii/LiDZQL26}, which limits transfer to unseen categories. Open-vocabulary (OV) methods alleviate this by lifting 2D foundation-model features into 3D representations for relational reasoning over novel concepts~\cite{koch2024open3dsg, zhang2023learning, chen2024clip}. Representative examples include ConceptGraphs~\cite{gu2024conceptgraphs}, Open3DSG~\cite{koch2024open3dsg}, and RelationField~\cite{koch2025relationfield}. However, these approaches mainly ground relations through vision--language similarity rather than explicit metric geometry. Since VLMs receive limited 3D spatial supervision during pre-training~\cite{ cheng2024spatialrgpt}, their spatial predictions remain unreliable under depth ambiguity and symmetric layouts. GHOST addresses this by grounding predicate prediction in explicit 3D geometry via discrete generative modeling.

\noindent\textbf{Discrete Diffusion for Structured Generation.}
Diffusion models have been extended from continuous generation~\cite{Ho2020, song2020denoising, rombach2022high,DBLP:conf/nips/LiGZLXOQ25,hu2026can,yin2025knowledge} to discrete state spaces. D3PM~\cite{austin2021structured} introduces structured categorical diffusion, VQ-Diffusion~\cite{Gu_2022_CVPR} applies it to vector-quantized image synthesis, and DiGress~\cite{vignac2023digress} models graph generation as iterative denoising over node and edge categories. Subsequent work improves structural fidelity with continuous-time formulations~\cite{xu2024discrete} and unified token-based generation~\cite{pan2025generative}. This paradigm is naturally suited to scene graphs, where both object labels and predicates are discrete and benefit from iterative refinement~\cite{tang2020unbiased, lin2020gps, yang2018graph}. GHOST is the first to apply hybrid diffusion to 3D scene graphs, jointly modeling geometry and predicate tokens with an intra-step geometry-to-relation denoising schedule for spatially consistent generation.

%% file: sec/4_method-ht.tex
\section{Preliminaries and Key Insights}
\label{sec:prelim}

\noindent\textbf{Problem formulation.}
We formulate 3D scene graph generation (3DSGG) as \emph{geometry-constrained graph generation}.
Given $N$ detected objects $\mathcal{V}=\{v_i\}_{i=1}^{N}$ from an off-the-shelf 3D detector, 
the goal is to predict a typed directed multigraph $\mathcal{G}=(\mathcal{V},\mathcal{E})$, where $\mathcal{E}\subseteq\mathcal{V}\times\mathcal{V}\times\mathcal{C}$ is the edge set and $\mathcal{C}$ is the set of all predicate classes.

Each detected object $v_i$ provides a category label $y_i\in\mathcal{Y}$ and an observed geometry vector $\boldsymbol{g}_i^{\mathrm{obs}}\in\mathbb{R}^{d_g}$, encoding the center, size, and orientation of its detector bounding box. We denote the fixed detector condition by $V=\{(y_i,\boldsymbol{g}_i^{\mathrm{obs}})\}_{i=1}^{N}$. Separately, $\boldsymbol{G}_t=[\boldsymbol{g}_{1,t};\dots;\boldsymbol{g}_{N,t}]\in\mathbb{R}^{N\times d_g}$ denotes the latent geometry variable evolved by diffusion, with $\boldsymbol{G}_T\sim\mathcal{N}(\boldsymbol{0},\boldsymbol{I})$. 

For relations, we consider all ordered pairs $\mathcal{P}=\{(i,j)\mid i\neq j\}$ and assign each pair $p\in\mathcal{P}$ a predicate token $z_E^{(p)}\in[M]$, where $[M]=\{1,\dots,M\}$ and $M$ is the predicate vocabulary size.
We write $\mathbf{Z}_E\in[M]^{|\mathcal{P}|}$ for the collection of all tokens.
The scene-graph state is thus factored into a continuous geometric component~$\mathbf{G}$ and a discrete semantic component~$\mathbf{Z}_E$.

\noindent\textbf{Spatial predicates as geometric constraints.}
A key observation is that spatial predicates are \emph{determined by geometry}: relations such as \emph{left of}, \emph{above}, and \emph{inside} each correspond to a composition of a small set of geometric primitives---axis comparisons, distance tests, and containment ratios.
We encode this through a language-grounded constraint function.
For an ordered pair $(i,j)\in\mathcal{P}$ with candidate predicate~$c$, we define
\begin{equation}
\label{eq:constraint_def}
\Phi(\mathbf{g}_i,\mathbf{g}_j;\,\mathbf{w}_c)\;\geq\;0
\;\iff\;
\text{predicate }c\text{ holds for }(i,j),
\end{equation}
where $\mathbf{g}_i,\mathbf{g}_j$ are the geometry vectors of nodes $v_i$ and $v_j$, and $\mathbf{w}_c\in\mathbb{R}^K$ is a weight vector \emph{predicted from the language embedding} of~$c$ via a learned mapping (Sec.~\ref{sec:method:guidance}).
Because $\mathbf{w}_c$ is derived from text, the constraint generalizes to predicates unseen during training.

\noindent\textbf{Joint estimation via diffusion.}
The constraint formulation implies that geometry and predicates should be resolved \emph{jointly}: a one-shot predictor commits to labels before the 3D layout is clear, and post-hoc correction can detect conflicts but cannot recover from them.
We therefore maintain a joint state $\mathcal{S}_t=(\mathbf{G}_t,\mathbf{Z}_{E,t})$, where subscript $t\in\{0,\dots,T\}$ indexes the diffusion timestep ($t{=}0$: clean; $t{=}T$: fully corrupted), and recover $\mathcal{S}_0$ through a learned reverse Markov chain with parameters~$\theta$:
\begin{equation} \label{eq:joint_reverse} p_\theta(\mathcal{S}_0\!\mid\!\mathcal{V},\mathbf{C}_{\text{ctx}}) \!=\!\int_{\mathbf{G}_{1:T}}\!\sum_{\mathbf{Z}_{E,1:T}} p(\mathcal{S}_T)\prod_{t=1}^{T} p_\theta(\mathcal{S}_{t-1}\!\mid\!\mathcal{S}_t,\mathcal{V},\mathbf{C}_{\text{ctx}}) \,d\mathbf{G}_{1:T}, \nonumber \end{equation}
where $\mathbf{G}_{1:T}=(\mathbf{G}_1,\dots,\mathbf{G}_T)$, the prior is $p(\mathcal{S}_T)=\mathcal{N}(\mathbf{G}_T;\mathbf{0},\mathbf{I})\cdot\mathrm{Uniform}(\mathbf{Z}_{E,T})$ (standard Gaussian for geometry, fully masked for predicates), and $\mathbf{C}_{\text{ctx}}\in\mathbb{R}^{d_c}$ is a global scene-condition vector (Sec.~\ref{sec:method:repr}).
Diffusion is well suited here because it maintains a joint distribution over geometry and predicates throughout inference, resolving both from global uncertainty to geometric consistency.

\noindent\textbf{Background.}
For continuous geometry, we adopt  DDPM~\cite{Ho2020}: the forward process adds Gaussian noise via
$q(\mathbf{g}_t\mid\mathbf{g}_{t-1})=\mathcal{N}(\sqrt{\alpha_t}\,\mathbf{g}_{t-1}, $ $(1{-}\alpha_t)\mathbf{I})$,
yielding marginal $q(\mathbf{g}_t\mid\mathbf{g}_0)=\mathcal{N}(\sqrt{\bar\alpha_t}\,\mathbf{g}_0,(1{-}\bar\alpha_t)\mathbf{I})$, and the reverse kernel is
$p_\theta(\mathbf{g}_{t-1}\mid\mathbf{g}_t)=\mathcal{N}(\boldsymbol{\mu}_\theta(\mathbf{g}_t,t),\tilde\beta_t\mathbf{I})$.
For discrete predicates, we adopt D3PM~\cite{austin2021structured}: each token evolves under a row-stochastic transition matrix~$Q_t$ that progressively masks tokens into an absorbing \texttt{[MASK]} state, with training posterior
$q(z_{t-1}\mid z_t,z_0)\propto(z_tQ_t^\top)\odot(z_0\bar Q_{t-1})$.

\begin{figure*}[t]
  \centering
  \includegraphics[width=1\textwidth]{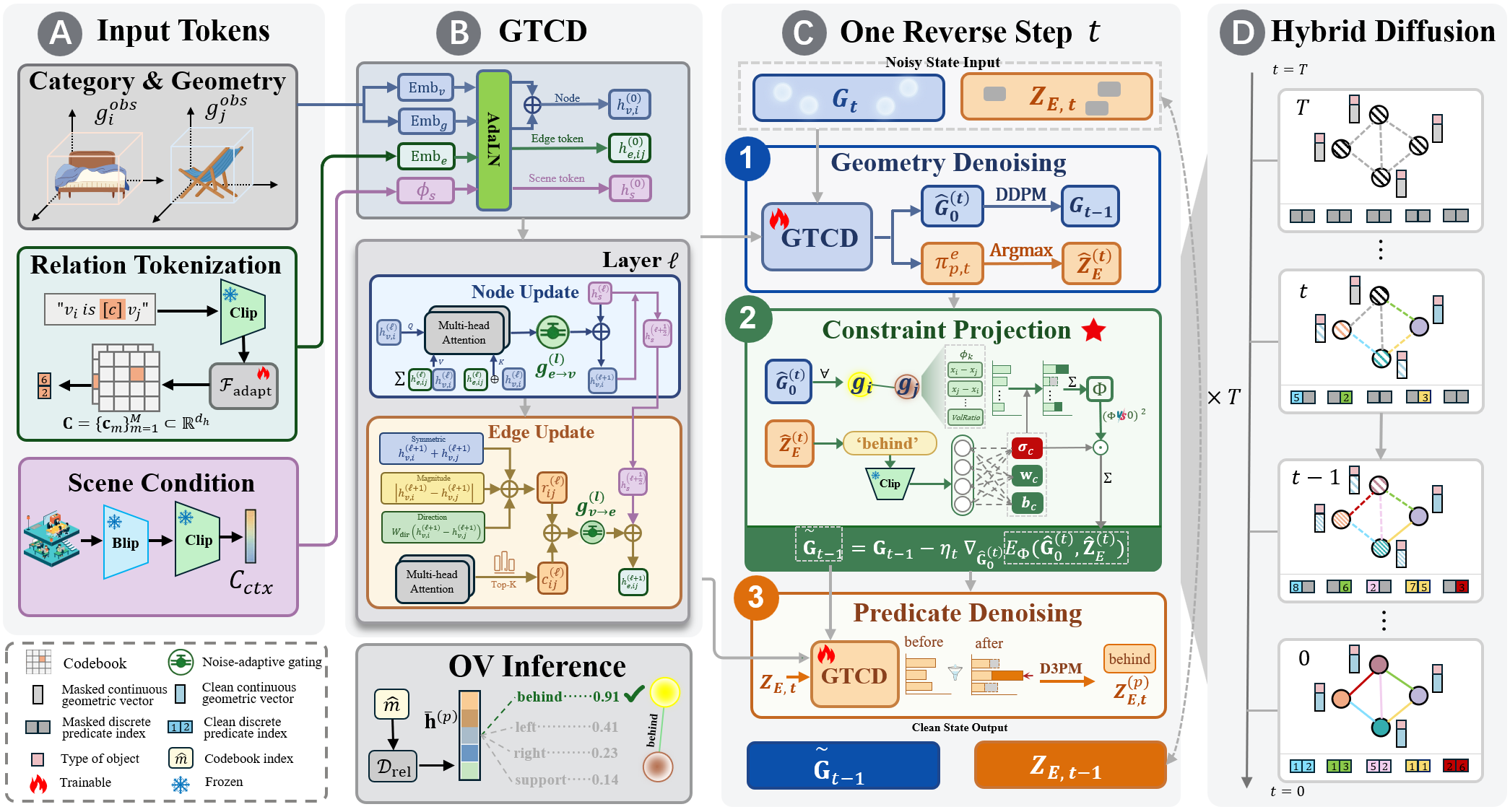}
\caption{\textbf{Overview of GHOST.}
GHOST jointly denoises object geometry and relation predicates through a hybrid reverse diffusion process. At each step, the Geometry-Topology Collaborative Denoiser (GTCD) first refines object geometry, then applies language-grounded constraint guidance to enforce geometric consistency, and finally decodes predicate tokens conditioned on the corrected layout. This denoise → project → decode ordering ensures that the recovered scene graph is both spatially coherent and open-vocabulary capable.}
  \label{fig:main}
\end{figure*}

\section{Method}
\label{sec:method}

GHOST casts 3DSGG as \emph{geometry-constrained relational decoding}, which recovers a 3D layout through diffusion, then decodes predicates from that layout under explicit geometric constraints, rather than classifying relations from visual--linguistic features in a single pass.
First, we tokenize predicate semantics into a discrete codebook so that both geometry and predicates live in spaces amenable to diffusion (\ref{sec:method:repr}).
Second, we define a hybrid reverse process in which geometry is denoised first and predicates are decoded conditioned on the recovered layout, with a language-grounded constraint mechanism (\ref{sec:method:diffusion}--\ref{sec:method:guidance}).
Third, we design a Geometry-Topology Collaborative Denoiser that  couples node and edge reasoning (\ref{sec:method:arch}).

\smallskip
\noindent\textbf{Row-level notation convention.}
Throughout this section, bold uppercase letters ($\mathbf{G}_t$, $\hat{\mathbf{G}}_0^{(t)}$, $\tilde{\mathbf{G}}_{t-1}$) denote $N{\times}d_g$ geometry matrices at diffusion step~$t$, and bold lowercase letters with a node subscript ($\mathbf{g}_{i,t}$, $\hat{\mathbf{g}}_{i,0}^{(t)}$, $\tilde{\mathbf{g}}_{i,t-1}$) denote their $i$-th row, i.e.\ the geometry of object~$v_i$.
We write $\mathbf{p}_i\in\mathbb{R}^3$ for the 3D center extracted from~$\mathbf{g}_i$. Throughout, $\mathbf{g}_i^{\mathrm{obs}}$ denotes the fixed detector geometry, whereas $\mathbf{g}_{i,t}$ denotes the latent geometry at diffusion step $t$.

\subsection{Scene Graph Tokenization}
\label{sec:method:repr}

\textbf{Geometry.}
The detector geometry ($\mathbf{g}_{i}^{\mathrm{obs}}$) is used as fixed conditioning, while the latent geometry ($\mathbf{g}_{i,t}$) is maintained to preserve metric precision and can be directly evaluated by the constraint function in Eq.~\ref{eq:constraint_def}.

\noindent\textbf{Relation tokenization.}
Edge predicates occupy a high-dimensional open semantic space that is ill-suited to direct diffusion; we therefore compress them into discrete tokens.
For each ordered pair $p=(i,j)$ with ground-truth predicate~$c$, we build a text template ``$y_i$ is $c$ $y_j$'' and extract a CLIP embedding $\mathbf{u}^{(p)}\in\mathbb{R}^{d_e}$, where $d_e$ is the CLIP feature dimension.
A lightweight residual adapter with learnable scalar~$\gamma\in\mathbb{R}$ is employed:
\begin{equation}
\label{eq:adapter}
\tilde{\mathbf{u}}^{(p)}
=\mathbf{u}^{(p)}+\gamma\cdot\mathcal{F}_{\text{adapt}}\bigl(\mathrm{LN}(\mathbf{u}^{(p)})\bigr),
\end{equation}
where $\mathcal{F}_{\text{adapt}}$ is a two-layer MLP and $\mathrm{LN}$ is layer normalization.
A learned encoder $\mathcal{E}_{\text{rel}}:\mathbb{R}^{d_e}\to\mathbb{R}^{d_h}$ projects the result into a latent space, which is vector-quantized against a codebook $\mathbf{C}=\{\mathbf{c}_m\}_{m=1}^M\subset\mathbb{R}^{d_h}$:
\begin{equation}
\hat m=\arg\min_{m\in[M]}\bigl\|\mathcal{E}_{\text{rel}}(\tilde{\mathbf{u}}^{(p)})-\mathbf{c}_m\bigr\|_2^2,
\qquad
z_E^{(p)}=\hat m.
\end{equation}
The quantizer is pretrained offline with a standard VQ-VAE objective~\cite{van2017neural,he2020sneq}; CLIP remains frozen throughout.
At inference, a symmetric decoder $\mathcal{D}_{\text{rel}}:\mathbb{R}^{d_h}\to\mathbb{R}^{d_e}$ maps codebook vectors back to the CLIP embedding space for open-vocabulary scoring.

\noindent\textbf{Scene condition.}
An image captioning model~\cite{li2022blip,li2023blip} applied to multi-view renderings produces a text summary of the global scene layout.
This caption is CLIP-encoded and projected through a small MLP to yield a condition vector~$\mathbf{C}_{\text{ctx}}\in\mathbb{R}^{d_c}$, injected into the denoiser at every reverse step.

\subsection{Hybrid Diffusion with Geometric Constraint}
\label{sec:method:diffusion}

The clean state is $\mathcal{S}_0=(\mathbf{G}_0,\mathbf{Z}_{E,0})$.
We adopt a \emph{forward-decoupled, reverse-coupled} design, where the two modalities are corrupted independently to avoid cross-modal noise interference during training, but recovered jointly under geometric constraints at inference.

\noindent\textbf{Forward process.}
The joint corruption factorizes as
$q(\mathcal{S}_{1:T}\mid\mathcal{S}_0)
=q_g(\mathbf{G}_{1:T}\mid\mathbf{G}_0)\cdot q_e(\mathbf{Z}_{E,1:T}\mid\mathbf{Z}_{E,0})$,
where geometry follows Gaussian diffusion \cite{Ho2020,sohl2015deep} and predicates  categorical masking as Sec.~\ref{sec:prelim}.

\noindent\textbf{Reverse process.}
At each step~$t$, the reverse transition factorizes sequentially:
\begin{equation}
\label{eq:reverse_factor}
\begin{split}
p_\theta(\mathcal{S}_{t-1}\mid\mathcal{S}_t,\mathcal{V},\mathbf{C}_{\text{ctx}})
\;=\;\;
&p_\theta^g(\mathbf{G}_{t-1}\mid\mathbf{G}_t,\mathbf{Z}_{E,t},\mathcal{V},\mathbf{C}_{\text{ctx}})\\
\cdot\;\;
&p_\theta^e(\mathbf{Z}_{E,t-1}\mid\tilde{\mathbf{G}}_{t-1},\mathbf{Z}_{E,t},\mathcal{V},\mathbf{C}_{\text{ctx}}),
\end{split}
\end{equation}
where $\tilde{\mathbf{G}}_{t-1}$ is the geometry \emph{after} the constraint projection defined below.
This factorization encodes a structural prior: predicate inference is always conditioned on a geometrically consistent layout.
We now describe each sub-step.

\emph{Step~1: Geometry denoising.}
The denoiser predicts a clean geometry estimate
$\hat{\mathbf{G}}_0^{(t)}=f_\theta^g(\mathbf{G}_t,\mathbf{Z}_{E,t},\mathcal{V},\mathbf{C}_{\text{ctx}},t)$,
and a new sample is drawn from the DDPM posterior:
\begin{equation}
\label{eq:geo_sample}
\mathbf{G}_{t-1}\sim\mathcal{N}\bigl(
\tilde{\boldsymbol{\mu}}_t(\mathbf{G}_t,\hat{\mathbf{G}}_0^{(t)}),\;
\tilde\beta_t\mathbf{I}\bigr),
\end{equation}
where $\tilde{\boldsymbol{\mu}}_t$ is the standard DDPM posterior mean~\cite{Ho2020}.
Note that the geometry branch conditions on the current predicate state~$\mathbf{Z}_{E,t}$, allowing progressively clarified relational structure to backward-correct geometric estimates.

\emph{Step~2: Constraint projection.}
The freshly sampled~$\mathbf{G}_{t-1}$ is not guaranteed to satisfy the geometric semantics of the predicted predicates.
We bridge this gap with a language-grounded constraint guidance mechanism (Sec.~\ref{sec:method:guidance}).
Let $\hat{\mathbf{Z}}_E^{(t)}=\{\hat c_{ij}^{(t)}\}_{(i,j)\in\mathcal{P}}$ collect the current best predicate for each edge, where $\hat c_{ij}^{(t)}=\arg\max_m\boldsymbol{\pi}_{(i,j),t}^\theta[m]$ and $\boldsymbol{\pi}_{(i,j),t}^\theta$ is the denoiser's predicted predicate distribution.
We correct the geometry with a gradient step on a differentiable constraint energy~$E_\Phi$ (Eq.~\ref{eq:constraint_energy}):
\begin{equation}
\label{eq:guidance_step}
\tilde{\mathbf{G}}_{t-1}
=\mathbf{G}_{t-1}
-\eta_t\,\nabla_{\hat{\mathbf{G}}_0^{(t)}} E_\Phi\bigl(\hat{\mathbf{G}}_0^{(t)},\hat{\mathbf{Z}}_E^{(t)}\bigr),
\end{equation}
where $\eta_t=\eta_{\max}\cos\bigl(\frac{\pi}{2}\cdot\frac{t}{T}\bigr)$ is a cosine guidance schedule that applies no correction at $t{=}T$ (when geometry is pure noise) and increases strength as estimates become reliable.

\emph{Step~3: Predicate denoising.}
The denoiser predicts a clean predicate distribution
$\boldsymbol{\pi}_{p,t}^\theta=f_\theta^e(\tilde{\mathbf{G}}_{t-1},\mathbf{Z}_{E,t},\mathcal{V},\mathbf{C}_{\text{ctx}},t)\in\Delta^{M-1}$
for each edge~$p=(i,j)$.
Before sampling, the distribution is reweighted to suppress spatially inconsistent candidates (Eq.~\ref{eq:pred_reweight}).
The token is then drawn from the D3PM posterior:
\begin{equation}
\label{eq:pred_sample}
z_{E,t-1}^{(p)}\sim\mathrm{Cat}\Bigl(
\mathbf{p}_{p,t}\propto
(z_{E,t}^{(p)}Q_t^\top)\odot(\tilde{\boldsymbol{\pi}}_{p,t}^\theta\,\bar Q_{t-1})
\Bigr).
\end{equation}
Because every step follows this strict \emph{denoise}$\,\to\,$\emph{project}$\,\to\,$\emph{decode} ordering, the diffusion trajectory progressively tightens the joint distribution over geometry and predicates.

\subsection{Language-Grounded Constraint Guidance}
\label{sec:method:guidance}

We now introduce the constraint energy~$E_\Phi$ and predicate reweighting used in the reverse process above.
The design goal is threefold: (i)~constraints must be differentiable so they can guide geometry via gradients, (ii)~they must be language-conditioned so they generalize to unseen predicates, and (iii)~they must gracefully ignore non-spatial predicates.

\noindent\textbf{Geometric primitive basis.}
We define a fixed set of $K$ differentiable primitives $\{\phi_k\}_{k=1}^K$ over pairs of geometry vectors $(\mathbf{g}_i,\mathbf{g}_j)$:
\begin{equation}
\label{eq:primitives}
\begin{aligned}
&\phi_1=x_j-x_i,\quad \phi_2=x_i-x_j, 
&\phi_3=y_j-y_i,\quad \phi_4=y_i-y_j, \\
&\phi_5=z_j-z_i,\quad \phi_6=z_i-z_j, 
&\phi_7=\|\mathbf{p}_j-\mathbf{p}_i\|_2-\tau_{\text{near}}, \\
&\phi_8=\mathrm{IoU}_{3D}(\mathbf{g}_i,\mathbf{g}_j)-\tau_{\text{contain}}, 
&\phi_9=\mathrm{VolRatio}(\mathbf{g}_i,\mathbf{g}_j)-1, %
\end{aligned}
\end{equation}
where $\mathbf{p}_i,\mathbf{p}_j\in\mathbb{R}^3$ are the 3D box centers extracted from $\mathbf{g}_i,\mathbf{g}_j$, and $\tau_{\text{near}},\tau_{\text{contain}}\in\mathbb{R}$ are learnable thresholds.
These primitives are predicate-agnostic; they form a shared geometric vocabulary from which any spatial constraint can be composed.

\noindent\textbf{From language to constraints.}
Given any predicate candidate~$c$ (seen or unseen), we extract its CLIP-adapted embedding~$\tilde{\mathbf{u}}_c\in\mathbb{R}^{d_e}$ using the same adapter as Eq.~\ref{eq:adapter}, and map it to a composition weight vector, a bias, and a spatiality score via three lightweight heads with shared parameters~$\omega$:
\(
\label{eq:constraint_gen}
\mathbf{w}_c=\mathcal{G}_\omega(\tilde{\mathbf{u}}_c)\in\mathbb{R}^K,
~
b_c=\mathcal{B}_\omega(\tilde{\mathbf{u}}_c)\in\mathbb{R},
~
\sigma_c=\sigma\bigl(\mathcal{H}_\omega(\tilde{\mathbf{u}}_c)\bigr),
\)
where $\mathcal{G}_\omega,\mathcal{B}_\omega$ are two-layer MLPs, $\mathcal{H}_\omega$ is a single linear layer, and $\sigma(\cdot)$ is the sigmoid function.
The scalar $\sigma_c$ is a spatiality gate which predicts how spatial predicate~$c$ is, ensuring that non-spatial predicates (e.g., \emph{made of}, \emph{has}) receive near-zero guidance.
The composed constraint for predicate~$c$ on pair $(i,j)$ is
\(
\label{eq:composed_constraint}
\Phi(\mathbf{g}_i,\mathbf{g}_j;\,\mathbf{w}_c,b_c)
=\textstyle\sum_{k=1}^K w_{c,k}\,\phi_k(\mathbf{g}_i,\mathbf{g}_j)+b_c.
\)
Because $\tilde{\mathbf{u}}_c\mapsto\mathbf{w}_c$ is smooth, semantically similar predicates (e.g., ``left of'' and ``to the west of'') receive similar weights, enabling open-vocabulary generalization.

\noindent\textbf{Constraint energy.}
Using the notation from Sec.~\ref{sec:method:diffusion}, the global constraint energy at step~$t$ is
\begin{equation}
\label{eq:constraint_energy}
E_\Phi\bigl(\hat{\mathbf{G}}_0^{(t)},\hat{\mathbf{Z}}_E^{(t)}\bigr)
\!=\!\!\!\!\sum_{(i,j)\in\mathcal{P}}
\!\!\!\sigma_{\hat c_{ij}^{(t)}}
\max\Bigl(0,-\Phi\bigl(\hat{\mathbf{g}}_{i,0}^{(t)},\hat{\mathbf{g}}_{j,0}^{(t)};\!\mathbf{w}_{\hat c_{ij}^{(t)}},b_{\hat c_{ij}^{(t)}}\bigr)\Bigr)^{2},
\end{equation}
where $\hat{\mathbf{g}}_{i,0}^{(t)}$ is the $i$-th row of the predicted clean geometry $\hat{\mathbf{G}}_0^{(t)}$.
The energy is zero when every predicted spatial predicate is geometrically satisfied and positive otherwise.
The gradient step in Eq.~\ref{eq:guidance_step} can be seen as classifier guidance~\cite{dhariwal2021diffusion} with the classifier replaced by an  language-grounded constraint energy.

After the correction, we reweight each predicate $m$ for edge $p{=}(i,j)$ by its constraint satisfaction on the corrected geometry:
\begin{equation}
\label{eq:pred_reweight}
\tilde{\boldsymbol{\pi}}_{p,t}^\theta[m]
\!\propto\!
\boldsymbol{\pi}_{p,t}^\theta[m]
\!\cdot\!\exp\Bigl(\!
{-}\lambda_t\,\sigma_m\!\max\bigl(0,-\Phi(\tilde{\mathbf{g}}_{i,t-1},\tilde{\mathbf{g}}_{j,t-1};\mathbf{w}_m,b_m)\bigr)^2
\Bigr),
\end{equation}
where $\tilde{\mathbf{g}}_{i,t-1}$ is the $i$-th row of $\tilde{\mathbf{G}}_{t-1}$, and $\lambda_t=\lambda_{\max}(1-t/T)$ is a linearly annealed temperature that applies strong reweighting at low noise levels.
The spatiality gate~$\sigma_m$ ensures that non-spatial candidates ($\sigma_m\approx 0$) are unaffected.

\subsection{Geometry-Topology Collaborative Denoiser}
\label{sec:method:arch}

The denoising functions $f_\theta^g$ and $f_\theta^e$ share a single backbone, i.e., \textbf{Geometry-Topology Collaborative Denoiser} (GTCD).
Its design is driven by two observations: (i)~node geometry should absorb relational evidence from edges, and (ii)~edge representations should be sensitive to the \emph{direction} between endpoints, not just their identity.

First, at step~$t$, node and edge features are initialized as

\begin{equation}
h^{(0)}_{v,i}
=
\operatorname{Emb}_{v}(y_i)
\oplus
\operatorname{Emb}_{\mathrm{obs}}(\mathbf{g}_i^{\mathrm{obs}})
\oplus
\operatorname{Emb}_{g}(\mathbf{g}_{i,t}), 
h^{(0)}_{e,ij}
=
\operatorname{Emb}_{e}\!\left(z^{(p)}_{E,t}\right).
\end{equation}
where $\oplus$ denotes concatenation; $\mathrm{Emb}_v$ and
$\mathrm{Emb}_e$ are learned lookup tables, while
$\mathrm{Emb}_{\mathrm{obs}}$ and $\mathrm{Emb}_g$ are separate
linear projections for the observed detector geometry and the latent
diffusion geometry, respectively. Here, $y_i$ and
$\mathbf{g}_i^{\mathrm{obs}}$ remain fixed across all reverse steps,
whereas $\mathbf{g}_{i,t}$ is iteratively denoised and
constraint-corrected.
All features are conditioned on the timestep via AdaLN~\cite{peebles2023scalable} with embedding $t_{\text{emb}}=\phi(t)\in\mathbb{R}^{d_t}$.
A global scene token $h_s^{(0)}=\phi_s(\mathbf{C}_{\text{ctx}})\in\mathbb{R}^d$ provides macroscopic context throughout all $L$ layers.

\noindent\textbf{Noise-adaptive gating.}
Edge tokens are unreliable at high noise levels.
To prevent noisy topology from corrupting geometric estimates early on, we gate all cross-branch messages with timestep-dependent scalars:
\begin{equation}
g_{e\to v}^{(\ell)}
=\sigma\bigl(W_{ev}[t_{\text{emb}}\oplus h_s^{(\ell)}]\bigr),
~
g_{v\to e}^{(\ell)}
=\sigma\bigl(W_{ve}[t_{\text{emb}}\oplus h_s^{(\ell+\frac12)}]\bigr),
\end{equation}
where $W_{ev},W_{ve}$ are learned linear projections and $h_s^{(\ell+\frac12)}$ is the scene token refreshed after the node update (defined below).
As denoising progresses ($t\to 0$), the gates open and geometry--topology interaction strengthens.

\noindent\textbf{Node update via edge-conditioned attention.}
Each node aggregates structural evidence from incident edges, allowing clarifying relational topology to correct geometric estimates.
Node~$v_i$ attends over its neighborhood $\mathcal{N}(i)=\{j:(i,j)\in\mathcal{P}\}$:
\begin{equation}
m_i^{(\ell)}
=\sum_{j\in\mathcal{N}(i)}\alpha_{ij}^{(\ell)}\,
\phi_{\text{inc}}\bigl(h_{e,ij}^{(\ell)},h_{v,j}^{(\ell)}\bigr),
\end{equation}
where $\alpha_{ij}^{(\ell)}=\mathrm{Softmax}_j\bigl(\frac{q_i^\top k_{ij}}{\sqrt{d}}\bigr)$ with $q_i=W_q h_{v,i}^{(\ell)}$, $k_{ij}=W_k(h_{e,ij}^{(\ell)}\oplus h_{v,j}^{(\ell)})$, and $\phi_{\text{inc}}$ is a two-layer MLP.
The gated node update is
\begin{equation}
\bar h_{v,i}^{(\ell+1)}=h_{v,i}^{(\ell)}+g_{e\to v}^{(\ell)}\odot m_i^{(\ell)},
\qquad
h_{v,i}^{(\ell+1)}=\phi_v\bigl(\bar h_{v,i}^{(\ell+1)}\oplus h_s^{(\ell)}\bigr),
\end{equation}
where $\phi_v$ is an MLP.
The scene token is refreshed by mean-pooling: $h_s^{(\ell+\frac12)}=\phi_s^v\bigl(h_s^{(\ell)},\frac{1}{N}\sum_i h_{v,i}^{(\ell+1)}\bigr)$.

\noindent\textbf{Edge update with orientation decomposition.}
For a directed edge $(i,j)$, we decompose the endpoint representation into order-invariant and direction-sensitive parts:
\begin{equation}
\label{eq:edge_decomp}
\boldsymbol{r}_{ij}^{(\ell)}
=\phi_{\text{pair}}\Bigl(
\underbrace{(h_{v,i}^{(\ell+1)}+h_{v,j}^{(\ell+1)})}_{\text{symmetric}}
\oplus
\underbrace{|h_{v,i}^{(\ell+1)}-h_{v,j}^{(\ell+1)}|}_{\text{magnitude}}
\oplus
\underbrace{\delta_{ij}^{\text{dir}}}_{\text{direction}}
\Bigr),
\end{equation}
where $\delta_{ij}^{\text{dir}}=W_{\text{dir}}(h_{v,i}^{(\ell+1)}-h_{v,j}^{(\ell+1)})$ is a learned directional projection.
The symmetric sum captures shared properties (proximity, co-occurrence); the signed difference and its magnitude encode the directional geometry needed for asymmetric predicates like \emph{left of} and \emph{above}.
To encourage consistency among edges that share an endpoint, each edge also attends to a Top-$K$ sparse neighborhood $\Omega_{ij}^{(\ell)}\subset\mathcal{N}_e(i,j)$ selected by attention score:
\(
c_{ij}^{(\ell)}
\!=\!
\textstyle\sum_{(u,v)\in\Omega_{ij}^{(\ell)}}
\beta_{ij,uv}^{(\ell)}\phi_{\mathrm{edge}}(h_{e,uv}^{(\ell)}),
\)
where $\beta_{ij,uv}^{(\ell)}$ are normalized scores and $\phi_{\text{edge}}$ is a linear projection.
The gated edge update is
$\bar h_{e,ij}^{(\ell+1)}\!=\!h_{e,ij}^{(\ell)}\!+\!g_{v\to e}^{(\ell)}\!\odot\!\phi_e(\boldsymbol{r}_{ij}^{(\ell)}\!\oplus \!c_{ij}^{(\ell)})$,
followed by scene-context injection $h_{e,ij}^{(\ell+1)}\!=\!\phi_e^{\text{glb}}(\bar h_{e,ij}^{(\ell+1)}\!\oplus\! h_s^{(\ell+\frac12)})$ and scene-token refresh.
After $L$ layers, the network outputs $(h_v^{(L)},h_e^{(L)})$ for the geometry and predicate prediction heads.

\subsection{Training and Open-Vocabulary Inference}
\label{sec:method:training}

We jointly train the denoiser, predicate tokenizer, and constraint generator with three objectives.
The \emph{geometry denoising loss} supervises clean geometry reconstruction and encourages the predicted layout to satisfy the ground-truth spatial predicate:
\begin{equation}
\label{eq:geo_loss}
\mathcal{L}_{\text{geo}}
\!=\!
\mathbb{E}_{t,\mathcal{S}_0,\boldsymbol{\epsilon}}
\Bigl[\!
N^{-1}\sum_{i=1}^{N}
\|\mathbf{g}_{i,0}\!-\!\hat{\mathbf{g}}_{i,0}^{(t)}\|_2^2
\!+\!
\mu |\mathcal{P}|^{-1}
\sum_{p\in\mathcal{P}}\!
\sigma_{c_p}\!
\max(0,-\Phi_{c_p}^{(t)})^2
\Bigr]. \nonumber
\end{equation}
where $p=(i,j)\in \mathcal{P}$ and
\(
\Phi_{c_p}^{(t)}
=
\Phi\bigl(
\hat{\mathbf{g}}_{i,0}^{(t)},
\hat{\mathbf{g}}_{j,0}^{(t)};
\mathbf{w}_{c_p},b_{c_p}
\bigr).
\)
Here, $\mu$ balances reconstruction and constraint consistency. By integrating the constraint term into $\mathcal{L}_{\text{geo}}$, the denoiser and constraint generator are optimized jointly.

The \emph{relation denoising loss} matches the predicted reverse distribution to the analytic D3PM posterior:
\begin{equation}
\begin{aligned}
\mathcal{L}_{\mathrm{rel}}
= {} & |\mathcal{P}|^{-1}
\mathbb{E}_{t,S_0,\mathbf{Z}_{E,t}} \Bigl[
\sum_{p \in \mathcal{P}}
\mathcal{D}_{\mathrm{KL}}\bigl(
q_{\mathbf{e}}(z_{E,t-1}^{(p)} \mid z_{E,t}^{(p)}, z_{E,0}^{(p)})
\\
& \qquad\qquad\qquad\quad \| \,
p_{\theta}^{e}(\cdot \mid \mathbf{G}_{t-1}, \mathbf{Z}_{E,t}, \mathbf{S}_t, C_{\mathrm{ctx}})
\bigr)
\Bigr].
\end{aligned}
\end{equation}
The auxiliary {spatiality gate loss} trains $\sigma_c$ to distinguish spatial predicates from non-spatial ones:
\begin{equation}
\label{eq:gate_loss}
\mathcal{L}_{\text{gate}}
=
-\frac{1}{|\mathcal{C}|}
\sum_{c\in\mathcal{C}}
\left[
\mathbb{I}_{\{c\in\mathcal{C}_s\}}\log\sigma_c
+
\bigl(1-\mathbb{I}_{\{c\in\mathcal{C}_s\}}\bigr)\log(1-\sigma_c)
\right]
\end{equation}
where $\mathcal{C}$ is the set of training predicates and $\mathcal{C}_s\subset\mathcal{C}$ denotes the spatial subset.
The total loss is
\(
\label{eq:total_loss}
\mathcal{L}_{\text{total}}
=
\lambda_g\mathcal{L}_{\text{geo}}
+
\lambda_r\mathcal{L}_{\text{rel}}
+
\lambda_s\mathcal{L}_{\text{gate}}.
\)

\noindent\textbf{Open-vocabulary inference.}
At $t{=}0$, each edge~$p$ yields a codebook index~$\hat m$.
The pretrained decoder maps it back to the CLIP embedding space: $\bar{\mathbf{h}}^{(p)}=\mathcal{D}_{\text{rel}}(\mathbf{c}_{\hat m})\in\mathbb{R}^{d_e}$.
For any candidate predicate~$c'$---including categories unseen during training---we compute its adapted embedding $\tilde{\mathbf{u}}_{c'}$ via Eq.~\ref{eq:adapter} and score it by cosine similarity:
\begin{equation}
s(c')=\cos\bigl(\bar{\mathbf{h}}^{(p)},\;\tilde{\mathbf{u}}_{c'}\bigr).
\end{equation}
For spatial candidates, the constraint generator simultaneously infers~$\mathbf{w}_{c'}$ from the same CLIP embedding.
Because the constraint guidance has been applied throughout the reverse process (Eq.~\ref{eq:guidance_step}), the recovered geometry already satisfies the learned constraints for the top-ranked predicates, so the cosine ranking inherits geometric consistency without post-processing.

%% file: sec/5_experiments.tex
\begin{table}[t]
\centering
\caption{%
  Predicate classification and relationship triplet detection on 3DSSG under
  closed-set and open-vocabulary (OV) protocols.
  $^\dagger$~Results reproduced from released code.
  ``--''~Not reported / no public implementation.
  \textbf{Bold}~=~best; \underline{underline}~=~second best per section.
}
\label{tab:comparative experiment}
\resizebox{\linewidth}{!}
{%
\begin{tabular}{l|cc|cc}
\toprule
\multirow{2}{*}{\textbf{Method}}
  & \multicolumn{2}{c|}{\textbf{Pred.\ Cls.}}
  & \multicolumn{2}{c}{\textbf{Triplets}} \\
\cmidrule(lr){2-3}\cmidrule(lr){4-5}
& R/mR@3 & R/mR@5 & R/mR@50 & R/mR@100 \\
\midrule
\multicolumn{5}{c}{\textit{Closed-set Setting}} \\
\midrule
3DSSG~\cite{wald2020learning}           & 0.89/0.54 & 0.93/0.56 & 0.40/0.26 & 0.66/0.42 \\
SGFN~\cite{wu2021scenegraphfusion}      & 0.97/0.68 & 0.99/0.69 & 0.85/0.51 & 0.87/0.55 \\
Lang3DSG~\cite{koch2024lang3dsg}        & 0.96/0.62 & 0.99/0.65 & 0.87/0.52 & 0.89/0.53 \\
SGRec3D~\cite{koch2024sgrec3d}          & 0.96/0.58 & 0.99/0.59 & 0.89/0.54 & 0.91/0.56 \\
VL-SAT~\cite{wang2023vl}               & 0.98/0.60 & 0.99/0.62 & 0.90/0.56 & 0.93/0.57 \\
SGFormer~\cite{lv2024sgformer}          & 0.98/0.61 & 0.99/0.63 & 0.91/0.57 & 0.93/0.60 \\
3D-HetSGP~\cite{ma2024heterogeneous}   & 0.98/0.62 & 0.99/0.63 & 0.91/0.58 & 0.94/0.60 \\
FROSS~\cite{hou2025fross}               & 0.98/0.65 & 0.99/0.66 & 0.93/0.62 & 0.95/0.63 \\
SCRSSG~\cite{yeo2025statistical}        & 0.98/0.64 & 0.99/0.65 & 0.93/0.60 & 0.95/0.62 \\
OCRL~\cite{heo2026object}
  & \underline{0.99}/\underline{0.72}
  & \underline{0.99}/\underline{0.73}
  & \underline{0.95}/\underline{0.68}
  & \underline{0.95}/\underline{0.69} \\
\midrule
\textbf{GHOST (Ours)}
  & \textbf{0.99}/\textbf{0.75}
  & \textbf{0.99}/\textbf{0.76}
  & \textbf{0.97}/\textbf{0.74}
  & \textbf{0.97}/\textbf{0.75} \\
\midrule
\multicolumn{5}{c}{\textit{Open-Vocabulary (OV) Setting}} \\
\midrule
CCL-3DSGG~\cite{chen2024clip}
  & 0.42/0.22 & 0.48/0.26 & 0.38/0.20 & 0.41/0.22 \\
ConceptGraphs$^\dagger$~\cite{gu2024conceptgraphs}
  & 0.60/0.34 & 0.65/0.36 & 0.55/0.30 & 0.57/0.31 \\
Open3DSG$^\dagger$~\cite{koch2024open3dsg}
  & 0.63/0.37 & 0.68/0.40 & 0.64/0.37 & 0.65/0.38 \\
OSU-3DSG~\cite{fei2026open}
  & 0.68/0.41 & 0.74/0.45 & 0.65/0.40 & 0.67/0.42 \\
SpatialRGPT$^\dagger$~\cite{cheng2024spatialrgpt}
  & 0.73/0.48 & 0.78/0.51 & 0.70/0.45 & 0.71/0.45 \\
RelationField$^\dagger$~\cite{koch2025relationfield}
  & 0.76/0.51 & 0.82/0.53 & 0.72/0.51 & 0.73/0.51 \\
ReLaGS~\cite{xie2026relags}
  & \underline{0.79}/\underline{0.55}
  & \underline{0.87}/\underline{0.61}
  & -- & -- \\
\midrule
\textbf{GHOST (Ours)}
  & \textbf{0.83}/\textbf{0.58}
  & \textbf{0.92}/\textbf{0.66}
  & \textbf{0.79}/\textbf{0.57}
  & \textbf{0.80}/\textbf{0.57} \\
\bottomrule
\end{tabular}%
}
\end{table}
\section{Experiments}
\label{sec:experiments}
\subsection{Experimental Settings}
\noindent\textbf{Datasets.}
All experiments are conducted on 3DSSG~\cite{wald2020learning}, comprising $1.5$K indoor scenes ($4.1$K sub-scenes) with $160$ object categories and $27$ predicate types.

\noindent\textbf{Baselines.}
Under the \emph{closed-set} and \emph{open-vocabulary} (OV) settings, we compare with 10~\cite{wald2020learning,wu2021scenegraphfusion,koch2024lang3dsg,koch2024sgrec3d,wang2023vl,lv2024sgformer,ma2024heterogeneous,hou2025fross,yeo2025statistical,heo2026object} and 7~\cite{chen2024clip,gu2024conceptgraphs,koch2024open3dsg,fei2026open,cheng2024spatialrgpt,koch2025relationfield,xie2026relags} recent methods, respectively. SpatialRGPT is adapted to RGB-D SGG using 3D cues from RelationField. We exclude OpenFunGraph~\cite{zhang2025open} due to its incompatible graph formulation and report only overall ReLaGS~\cite{xie2026relags} results because its implementation is unavailable.
We evaluate Predicate Classification, given ground-truth boxes and labels, and Relationship Triplet Detection, where objects and predicates are jointly predicted, using R@$K$ and mR@$K$. For OV evaluation, one-third of predicates are held out as unseen, and results are averaged over three splits and three seeds.

\noindent\textbf{Implementation Details.}
GHOST is trained in PyTorch on four RTX~4090 GPUs with Adam ($\text{lr}{=}10^{-3}$, batch size 8) and loss weights $\lambda_{\text{g}}{:}\lambda_{\text{r}}{:}\lambda_{\text{s}}=0.4{:}0.6{:}0.1$. We use $T{=}150$ diffusion steps, $L{=}6$ GTCD layers, a 128-entry dual relation codebook, $K{=}9$ geometric primitives, $\eta_{\max}{=}0.1$, and $\lambda_{\max}{=}5.0$.

\subsection{Comparison with State-of-the-Art Models}
\label{subsec:Comparison}

\noindent\textbf{Overall results.}
Tab.~\ref{tab:comparative experiment} reports results under both settings.
Under the \emph{closed-set} setting, GHOST surpasses the strongest baseline
OCRL by +3 points on both mR@3 and mR@5 while matching it on R@K, indicating stronger long-tail relation recognition.
Under the \emph{OV} setting, GHOST sets new state-of-the-art on all reported metrics: it outperforms ReLaGS by +4/+5 points in predicate R@3/R@5, and surpasses RelationField by +7 points in both R@50 and R@100 for triplet detection (+6 points in mR).
These consistent gains across both settings suggest that GHOST's geometry-first diffusion modeling yields a stronger 3D scene graph prior rather than exploiting vocabulary-specific biases.

\begin{table}[t]
\centering
\setlength{\tabcolsep}{2pt}
\caption{%
  Spatial vs.\ non-spatial predicate recall (OV setting).
}
\label{tab:spatial results}
\resizebox{\columnwidth}{!}
{%
\begin{tabular}{@{}l|ccc|ccc@{}}
\toprule
\multirow{2}{*}{Method}
  & \multicolumn{3}{c}{Spatial}
  & \multicolumn{3}{c}{Non-Spatial} \\
\cmidrule(r){2-4}\cmidrule(l){5-7}
& R/mR@1 & R/mR@3 & R/mR@5
& R/mR@1 & R/mR@3 & R/mR@5 \\
\midrule
CG~\cite{gu2024conceptgraphs}
  & 0.08/0.04 & 0.31/0.17 & 0.35/0.19
  & 0.26/0.14 & 0.70/0.43 & 0.75/0.45 \\
O3DSG~\cite{koch2024open3dsg}
  & 0.11/0.06 & 0.35/0.20 & 0.41/0.24
  & 0.29/0.17 & 0.72/0.45 & 0.77/0.49 \\
SRGPT~\cite{cheng2024spatialrgpt}
  & 0.22/0.14 & 0.50/0.32 & 0.55/0.36
  & 0.39/0.26 & 0.81/0.54 & 0.86/0.60 \\
RF~\cite{koch2025relationfield}
  & 0.26/0.17 & 0.53/0.36 & 0.60/0.40
  & \textbf{0.42}/\textbf{0.29} & 0.83/0.57 & 0.90/0.61 \\
\midrule
\textbf{GHOST}
  & \textbf{0.43}/\textbf{0.30} & \textbf{0.84}/\textbf{0.60} & \textbf{0.93}/\textbf{0.65}
  & 0.41/0.28 & \textbf{0.83}/\textbf{0.58} & \textbf{0.92}/\textbf{0.63} \\
\bottomrule
\end{tabular}%
}
\end{table}

\noindent\textbf{Spatial predicate results.}
Tab.~\ref{tab:spatial results} reveals a fundamental limitation of four representative VLM-based methods: a large performance gap between spatial and non-spatial predicates.  
The strongest baseline RelationField achieves spatial R/mR@1 of only 0.26/0.17 versus 0.42/0.29 on non-spatial predicates---a near $2\times$ deficit that grows even larger for ConceptGraphs (0.08 vs.\ 0.26 at R@1).
This disparity indicates that VLM-driven methods fit semantic co-occurrence statistics rather than performing metric geometric reasoning.
GHOST closes this gap decisively: spatial R/mR@1 improves to \textbf{0.43/0.30} (+17 points over RelationField) and R/mR@5 to \textbf{0.93/0.65} (+33 points), while non-spatial R/mR@5 of 0.92/0.63 remains on par with the best baseline (0.90/0.61), confirming that geometric constraint guidance does not compromise semantic generalization.

\noindent\textbf{Seen vs. unseen predicates.}
Tab.~\ref{tab:open_vocab} shows that GHOST consistently outperforms RelationField on both seen and unseen predicates, with about an 8-point improvement on average. Importantly, the gains on unseen predicates are comparable to those on seen ones, indicating that the improvement is not simply due to better fitting of visible categories. Instead, the results suggest that GHOST learns transferable geometry--relation mappings and exhibits genuine open-vocabulary generalization.

\begin{figure}[t]
\centering
\includegraphics[width=0.45\textwidth]{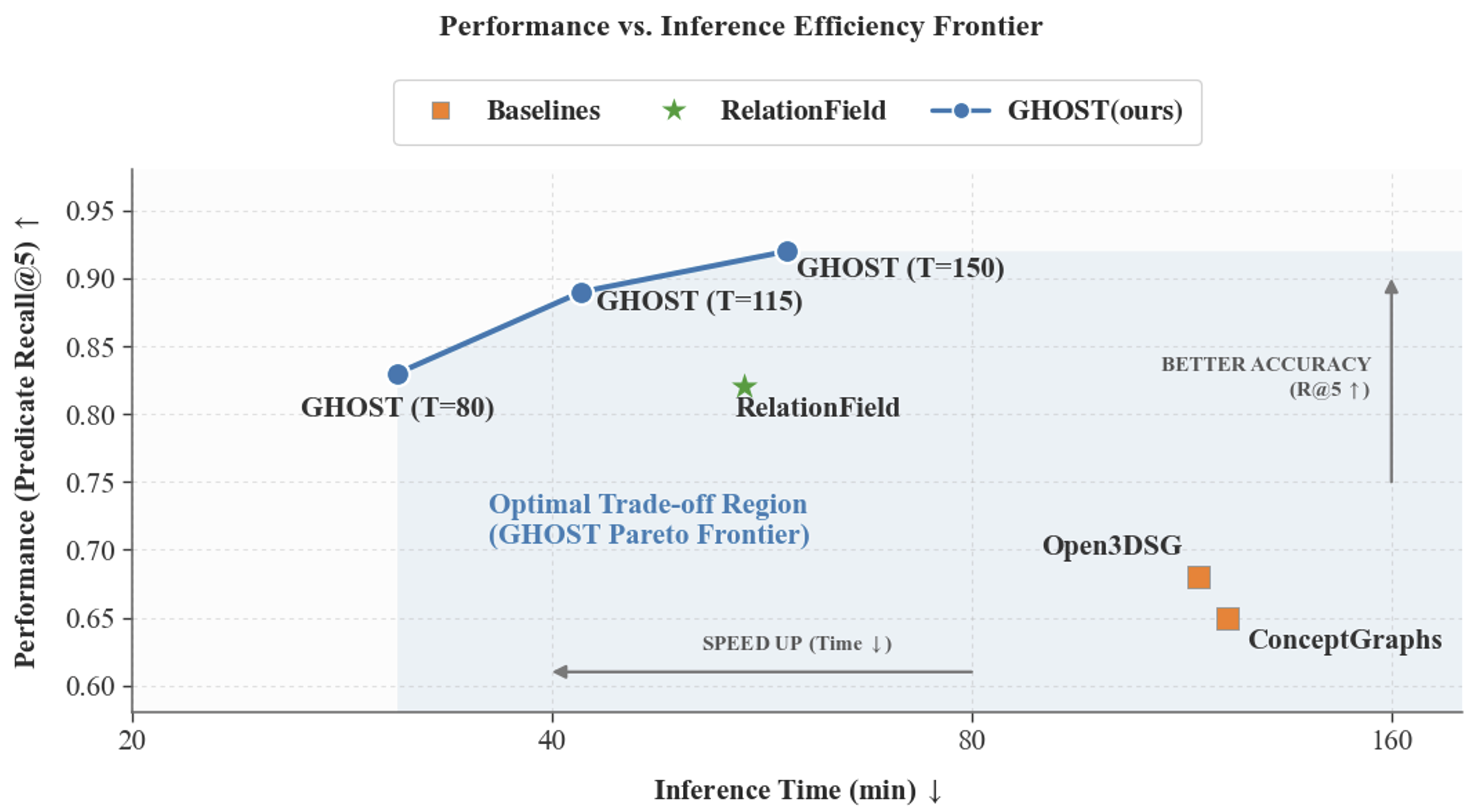}
\caption{%
  Accuracy--efficiency trade-off (predicate R@5).
  All GHOST variants share the same weights and environment and differ only in reverse-step count~$T$.
  ReLaGS is excluded due to the unavailable source code.
}
\label{fig:pe}
\end{figure}

\noindent\textbf{Inference efficiency.} As shown in Fig.~\ref{fig:pe}, GHOST defines a better accuracy--efficiency frontier than prior VLM-based pipelines. Even the fastest variant already surpasses RelationField with lower latency, while increasing $T$ further improves accuracy with only modest extra cost. This efficiency mainly comes from two designs: codebook-based relation discretization, which avoids expensive per-edge LLM inference, and noise-adaptive gating, which suppresses unnecessary cross-branch interactions during early denoising.

\begin{table}[t]
  \centering
  \small
  \caption{\textbf{Seen vs. unseen predicate recall (OV setting).}}
  \label{tab:open_vocab}
  \begin{tabular}{l | c c | c c}
    \toprule
    \multirow{2}{*}{\textbf{Method}} & \multicolumn{2}{c|}{\textbf{Seen Predicate}} & \multicolumn{2}{c}{\textbf{Unseen Predicate}} \\
    \cmidrule{2-5}
    & R/mR@3 & R/mR@5 & R/mR@3 & R/mR@5 \\
    \midrule
    ConceptGraphs ~\cite{gu2024conceptgraphs}  & 0.65/0.36 & 0.69/0.38 & 0.50/0.27 & 0.55/0.31 \\
    Open3DSG ~\cite{koch2024open3dsg}      & 0.67/0.40 & 0.72/0.43 & 0.56/0.33 & 0.61/0.36 \\
    SpatialRGPT ~\cite{cheng2024spatialrgpt}   & 0.77/0.50 & 0.80/0.54 & 0.67/0.44 & 0.74/0.48 \\
    RelationField ~\cite{koch2025relationfield} & \underline{0.79}/\underline{0.54} & \underline{0.85}/\underline{0.56} & \underline{0.73}/\underline{0.46} & \underline{0.78}/\underline{0.50} \\
    \midrule
    \textbf{GHOST (Ours)} & \textbf{0.85/0.60} & \textbf{0.94/0.67} & \textbf{0.81/0.55} & \textbf{0.87/0.63} \\
    \bottomrule
  \end{tabular}
\end{table}

\subsection{Ablation Study}
\label{subsec:ablation}

We ablate key design choices of GHOST on predicate R@$K$ and spatial R@$K$ (Tab.~\ref{tab:ablations study}).

\begin{table}[t]
\centering
\caption{Ablation of GHOST's components at the OV setting. Each variant removes or
replaces a single component while keeping all others fixed.}
\label{tab:ablations study}
\resizebox{\columnwidth}{!}
{%
\begin{tabular}{l|cc|cc}
\toprule
\multirow{2}{*}{\textbf{Variants}}
  & \multicolumn{2}{c|}{Predicate}
  & \multicolumn{2}{c}{Spatial Pred.} \\
\cmidrule(lr){2-3}\cmidrule(lr){4-5}
& R@3 & R@5 & R@1 & R@5 \\
\midrule
\multicolumn{5}{l}{\small\textit{Input configuration}} \\
w/ vision tokens          & 0.84 & 0.93 & 0.41 & 0.90 \\
w/o geometry tokens       & 0.61 & 0.64 & 0.12 & 0.32 \\
\midrule
\multicolumn{5}{l}{\small\textit{Annealing schedule}} \\
Pred.$\to$Geo.\ order & 0.60 & 0.63 & 0.13 & 0.33 \\
\midrule
\multicolumn{5}{l}{\small\textit{Diffusion mechanism}} \\
w/o diffusion (iterative refinement) & 0.59 & 0.63 & 0.14 & 0.35 \\
\midrule
\multicolumn{5}{l}{\small\textit{Constraint guidance}} \\
w/o guidance (both)                & 0.65 & 0.69 & 0.15 & 0.35 \\
w/o spatiality gate                & 0.75 & 0.80 & 0.38 & 0.87 \\
\midrule
\textbf{GHOST (full)}         & \textbf{0.83} & \textbf{0.92} & \textbf{0.43} & \textbf{0.93} \\
\bottomrule
\end{tabular}%
}
\end{table}

\noindent\textbf{Input configuration.}
Adding vision tokens yields only marginal gains ($+1$ point in predicate R@3/R@5) at roughly $2\times$ inference cost,
so we exclude them by default. Removing geometry tokens causes severe degradation (predicate R@3/R@5: 0.61/0.64; spatial R@1/R@5: 0.12/0.32), confirming that explicit geometry is essential for reliable spatial reasoning beyond label co-occurrence.

\noindent\textbf{Geometric annealing schedule.}
Reversing the intra-step order to predicate-first, geometry-second degrades predicate R@3/R@5 to 0.60/0.63 and spatial R@1/R@5 to 0.13/0.33. Predicting relations before resolving geometry encourages semantic shortcuts, whose errors subsequently bias geometry estimation.

\noindent\textbf{Diffusion mechanism.}
Replacing stochastic posterior sampling with deterministic updates causes a marked drop in Predicate R@3/5 (0.59/0.63) and Spatial R@1/5 (0.14/0.35), revealing a core weakness of purely iterative refinement: early commitment to an incorrect relation under sparse geometric evidence locks subsequent steps on a flawed trajectory. GHOST avoids this by maintaining a soft joint geometry–predicate distribution via diffusion, deferring consensus until sufficient evidence accumulates.

\noindent\textbf{Constraint guidance.}
We consider two ablated variants: \emph{w/o guidance (both)}, which removes both gradient geometry correction (Eq.~\ref{eq:guidance_step}) and predicate reweighting (Eq.~\ref{eq:pred_reweight}); and a variant that disables the spatiality gate $\sigma_c$.
Removing both components yields the sharpest overall performance drop (predicate R@3/R@5: 0.65/0.69; spatial R@1/R@5: 0.15/0.35), confirming that constraint guidance serves as the primary driver for geometric reasoning.
Removing the spatiality gate $\sigma_c$ significantly degrades overall performance (predicate R@3/R@5: 0.75/0.80) and spatial recall (spatial R@1/R@5: 0.38/0.87). This verifies that the gate protects purely semantic predicates from geometric gradient interference.

\begin{figure}[t]
  \centering
    \includegraphics[width=\linewidth]{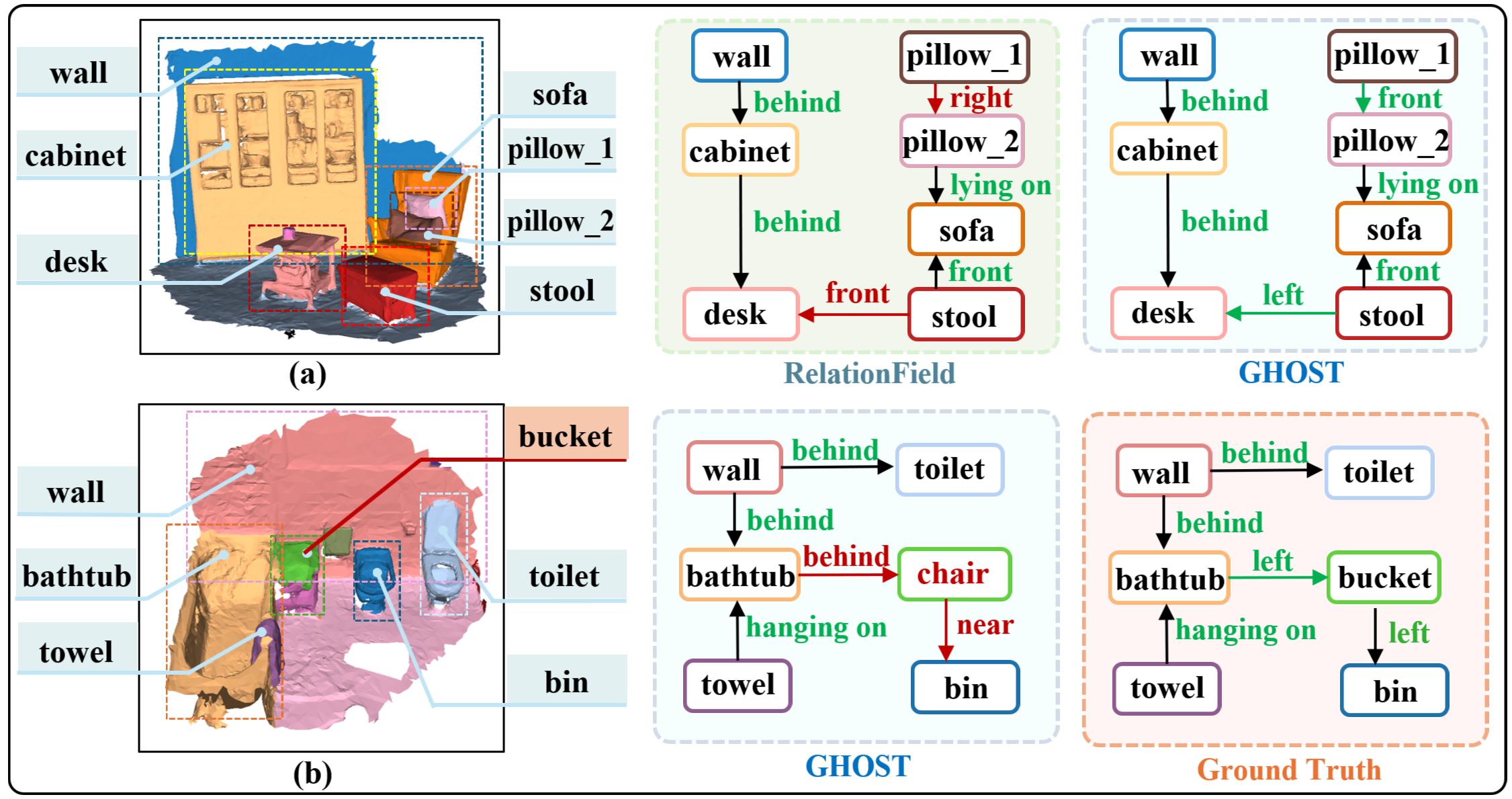}
    \label{fig:Qua2}
    \caption{Qualitative OV scene graph predictions on 3DSSG~\cite{wald2020learning}. Top: comparison between RelationField and GHOST. Bottom: a GHOST failure case caused by an object recognition error from the upstream detector. \textcolor{green}{Green} = correct; \textcolor{red}{red} = error.}
\label{fig:visualization_results}
\end{figure}

\noindent\textbf{Qualitative results.}
\label{sec:exp:vis}
Fig.~\ref{fig:visualization_results} presents an error mode corrected by GHOST and a representative failure case.
In scene~(a), the similar depths of \texttt{pillow\_1} and \texttt{pillow\_2} cause RelationField to confuse \emph{in front of} with \emph{right}, whereas GHOST preserves the correct 3D ordering through progressive constraint projection.
In scene~(b), the upstream detector misclassifies the bucket as a chair, and the incorrect object category consequently disrupts geometry--relation reasoning and propagates to relation prediction.

\begin{figure}[t] \centering \includegraphics[width=0.46\textwidth]{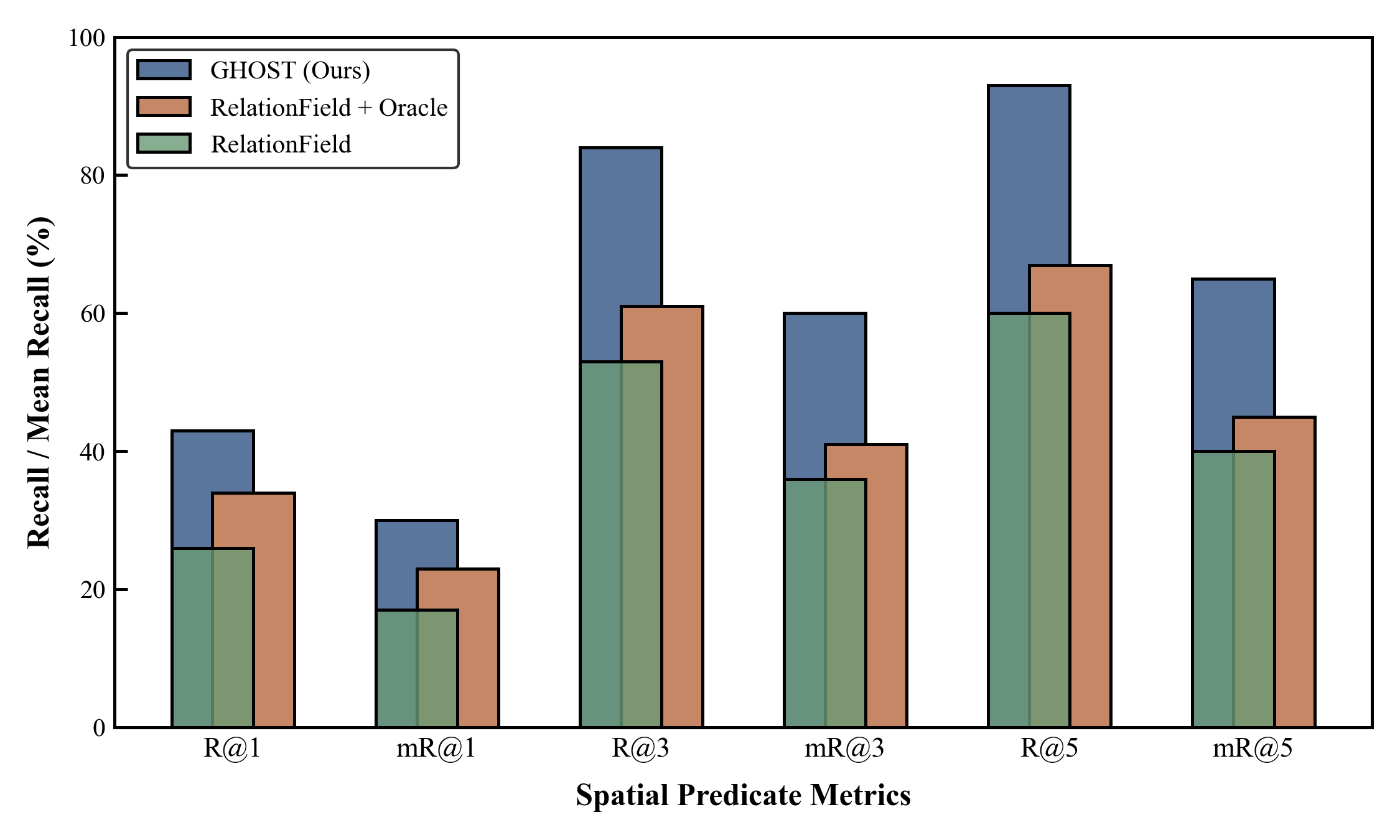}
\caption{Geometric Oracle Experiments. Even a perfect geometric oracle applied post-hoc yields only modest spatial gains over the one-shot baseline.}
\label{fig:oracle} 
\end{figure}

\subsection{Geometric Oracle Experiments}
\label{sec:exp:oracle}
VLM-based 3DSGG methods struggle with spatial predicates due to \textbf{co-occurrence bias}, which favors corpus statistics over metric geometry, and \textbf{one-shot commitment}, which fixes predictions before the scene layout is fully resolved. To assess whether post-hoc verification can address these issues, we augment RelationField~\cite{koch2025relationfield}, the strongest one-shot baseline, with a perfect geometric oracle that uses ground-truth bounding boxes to remove geometrically invalid candidates and re-rank the remainder by confidence.
As shown in Fig.~\ref{fig:oracle}, the oracle yields only modest improvements: spatial R/mR@1 increases from 0.26/0.17 to 0.34/0.23, R/mR@3 from 0.53/0.36 to 0.61/0.41, and R/mR@5 from 0.60/0.40 to 0.67/0.45. These results remain well below GHOST's 0.43/0.30, 0.84/0.60, and 0.93/0.65 without oracle access. Post-processing cannot recover correct predicates already excluded from the top-$K$ candidates by co-occurrence bias. In contrast, GHOST jointly maintains revisable distributions over geometry and predicates throughout reverse diffusion, delaying final decisions until sufficient geometric evidence has accumulated.

%% file: sec/6_Conclusions.tex
\section{Conclusion}
\label{sec:conclusion}
We presented GHOST, a hybrid diffusion framework that jointly denoises node geometry and edge predicates for open-vocabulary 3D scene graph generation. By integrating geometric annealing, language-grounded constraint guidance, and orientation-aware collaborative denoising, GHOST enables verifiable metric reasoning beyond co-occurrence-based VLM priors. It achieves state-of-the-art performance on 3DSSG in both closed-set and open-vocabulary settings, with particularly strong gains on spatial predicates.

\noindent\textbf{Limitations.}
The iterative reverse process introduces latency, although fewer diffusion steps provide a favorable efficiency--accuracy trade-off. Performance in complex scenes is also limited by the expressiveness of the hand-designed geometric primitives. Future work will explore learned primitives, part-level and dynamic scenes, and larger open-world benchmarks.

\section*{Acknowledgments}
This research was partially supported by the National Natural Science Foundation of China (NSFC)
(No. 62306064), Sichuan Science and Technology Program (granted No. 2024ZDZX0011 and No. 2025ZHCG0002) and Fundamental Research Funds for the Central Universities (No. ZYGX2025XJ049).